\documentclass[sigconf, nonacm]{acmart}

\AtBeginDocument{%
  }

\copyrightyear{2025}
\acmYear{2025}
\setcctype{cc-by}
\usepackage{colortbl}
\usepackage{tikz}
\usepackage{array}
\usepackage{multirow}
\usepackage{fontawesome}
\usepackage{array}
\usepackage{subcaption}
\newcolumntype{C}[1]{>{\centering}m{#1}}
\usepackage{CJKutf8}

\begin{document}

% \title[WebFAQ]{WebFAQ: Extracting Question and Answer pairs from the web}
\title[WebFAQ: A Multilingual Collection of Natural Q\&A Datasets for Dense Retrieval]{WebFAQ: A Multilingual Collection of Natural Q\&A Datasets for Dense Retrieval}

\author{Michael Dinzinger}
\email{michael.dinzinger@uni-passau.de}
% \authornote{Corresponding author}
\orcid{0009-0003-1747-5643}
\affiliation{%
  % \department{Chair of Data Science}
  \institution{University of Passau}
  \streetaddress{Innstraße 33}
  \city{Passau}
  \country{Germany}
  \postcode{94032}
}

\author{Laura Caspari}
\email{laura.caspari@uni-passau.de}
\orcid{0009-0002-6670-3211}
\affiliation{%
  % \department{Chair of Data Science}
  \institution{University of Passau}
  \streetaddress{Innstraße 33}
  \city{Passau}
  \country{Germany}
  \postcode{94032}
}

\author{Kanishka Ghosh Dastidar}
\email{kanishka.ghoshdastidar@uni-passau.de}
\orcid{0000-0003-4171-0597}
\affiliation{%
  % \department{Chair of Data Science}
  \institution{University of Passau}
  \streetaddress{Innstraße 33}
  \city{Passau}
  \country{Germany}
  \postcode{94032}
}

\author{Jelena Mitrovi\'{c}}
\email{jelena.mitrovic@uni-passau.de}
\orcid{0000-0003-3220-8749}
\affiliation{%
  % \department{Chair of Data Science}
  \institution{University of Passau}
  \streetaddress{Innstraße 33}
  \city{Passau}
  \country{Germany}
  \postcode{94032}
}

\author{Michael Granitzer}
\email{michael.granitzer@uni-passau.de}
\orcid{0000-0003-3566-5507}
\affiliation{%
  % \department{Chair of Data Science}
  \institution{University of Passau}
  \streetaddress{Innstraße 33}
  \city{Passau}
  \country{Germany}
  \postcode{94032}
}

% \affiliation{%

% \affiliation{%

\renewcommand{\shortauthors}{Dinzinger et al.}

\begin{abstract}
We present WebFAQ, a large-scale collection of open-do\-main question answering datasets derived from FAQ-style sche\-ma.org annotations. In total, the data collection consists of 96 million natural question-answer (QA) pairs across 75 languages, including 47 million (49\%) non-English samples.
WebFAQ further serves as the foundation for 20 monolingual retrieval benchmarks with a total size of 11.2 million QA pairs (5.9 million non-English). These datasets are carefully curated through refined filtering and near-duplicate detection, yielding high-quality resources for training and evaluating multilingual dense retrieval models.
To empirically confirm WebFAQ's efficacy, we use the collected QAs to fine-tu\-ne an in-domain pretrained XLM-RoBERTa model. Through this process of dataset-specific fine-tuning, the model achieves significant retrieval performance gains, which generalize -- beyond WebFAQ -- to other multilingual retrieval benchmarks evaluated in zero-shot setting.
Last but not least, we utilize WebFAQ to construct a set of QA-aligned bilingual corpora spanning over 1000 language pairs using state-of-the-art bitext mining and automa\-ted LLM-assessed translation evaluation. Due to our advanced, automated method of bitext dataset generation, the resulting bilingual corpora demonstrate higher translation quality compared to similar datasets.
WebFAQ and all associated resources are publicly available on Git\-Hub\footnote{\url{https://github.com/padas-lab-de/webfaq}} and HuggingFace.\footnote{\url{https://huggingface.co/PaDaS-Lab}}
\end{abstract}




% \received{05 February 2024}
% \received[revised]{-}
% \received[accepted]{-}

\maketitle

\section{Introduction}

\begin{figure}[t]
  \centering
  \vspace{0.5cm}
  \includegraphics[width=\linewidth]{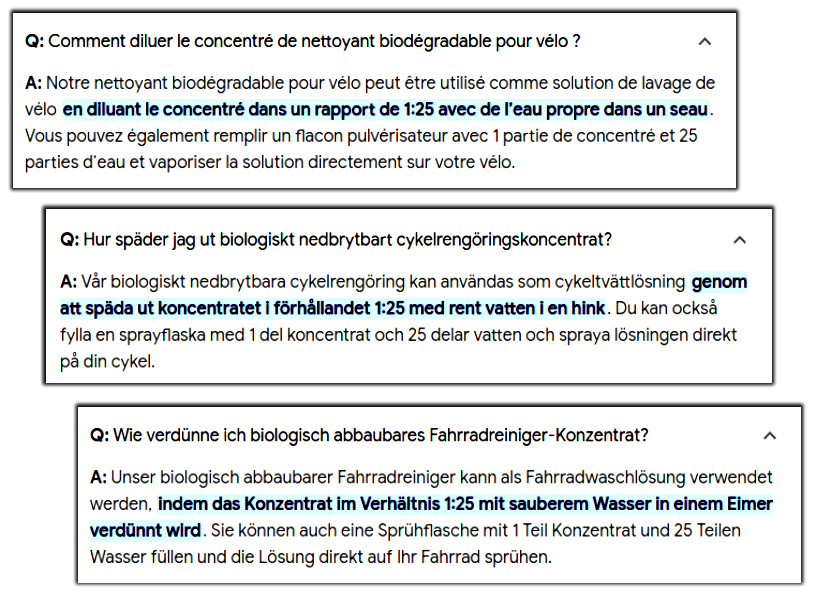}
  \caption{Exemplary FAQ entries across languages}
  \label{fig:examples}
  \Description{}
  % \vspace{-0.1cm}
\end{figure}

The rapid adoption of structured data annotations, such as Microdata and JSON-LD formats, have significantly transformed the way information is presented and consumed online. Among these structured resources, FAQ (Frequently Asked Questions) pages provide an exceptionally rich collection of natural question-answer (QA) pairs across numerous topics and languages (see Figure~\ref{fig:examples}). With the widespread use of schema.org annotations, search engines and web crawlers can extract these QA pairs efficiently, which provides a unique opportunity to leverage this data for research in the area of Natural Language Processing (NLP).

% \begin{figure}[t]
%   \begin{tabular}{l}
%   \toprule
%   \textbf{Q:} Como estas? (spa.) \\
%   \textbf{A:} Soy bien. \\
%   \midrule
%   \textbf{Q:} Wie geht es dir? (deu.) \\
%   \textbf{A:} Mir geht es gut. \\
%   \midrule
%   \textbf{Q:} How are you doing? (eng.) \\
%   \textbf{A:} I'm fine. \\
%   \bottomrule
%   \end{tabular}
%   \caption{WebFAQ example: Parallel Q\&A pairs}
%   \label{fig:examples}
%   \Description{}
% \end{figure}

% - Recent studies show that QA datasets can be successfully utilized for the pre-training of ODQA systems as well as retrieval models [Ref]. Such systems mean to accurately answer questions posed in natural language from an arbitrary domain, or retrieving a relevant answer from a large-scale corpus respectively.
% - Especially in this ``open-retrieval'' setting, a number of QA datasets have gained significant popularity in evaluating and fine-tuning language models (LM), such as MS MARCO [Ref], HotpotQA [Ref] and NQ [Ref]. These datasets are commonly used to evaluate embedding models, trained as bi-encoding dense retrievers, on the well-known Massive Text Embedding Benchmark (MTEB). Embedding models, such as ... and ..., have been fine-tuned on HotpotQA, etc. during their development, helping their overall retrieval performance.
% - Recent initiaves, such as Massive Multilingual Text Embedding Benchmark (MMTEB), a community-driven expansion of MTEB, demonstrate that there is an ongoing trend towards multilinguality and an increase in the diversity of evaluation tasks, including Question Answering Reranking, cross-lingual STS (Semantic Text Similarity) or cross-lingual retrieval. Consequently, multilingual and cross-lingual resources become more important.
% - With our work, we want to contribute the - to the best of our knowledge - first large-scale QA dataset with a focus on cross-linguality with natural questions that represents the entire diversity of the web and is not extracted from Wikipedia.

Furthermore, recent studies have demonstrated the effectiveness of Q\&A datasets in training Open-Domain Question Answering (ODQA) systems and retrieval models~\cite{Huber2022, Zhang2021, Campese2023, Lewis2021}. ODQA systems aim to accurately answer natural language questions a\-cross arbitrary domains, while retrieval models are designed to identify relevant documents or passages, and thus potential answers, from large-scale corpora. Particularly in this ``open-retriev\-al'' settings, various Q\&A benchmarks, such as MS MARCO~\cite{Bajaj2016-arxiv}, HotpotQA~\cite{Yang2018} and Natural Questions (NQ)~\cite{Kwiatkowski2019}, have become essential cornerstones for the current broad scientific progress in Neural IR. For instance, these datasets are included in the training of numerous embedding models as bi-enco\-ding dense retrievers~\cite{Wang2024-arxiv, Chen2024-M3, Zhang2024, Sturua2024-arxiv}, and are commonly used in the evaluation of MTEB (Massive Text Embedding Benchmark)~\cite{Muennighoff2023}.

Recent initiatives such as the Massive Multilingual Text Embedding Benchmark (MMTEB)~\cite{Enevoldsen2025-openreview}, an expansion of MTEB driven by the research community, demonstrate that there is an ongoing shift towards multilinguality and the diversification of evaluation tasks for text embedding models. The newly incorporated tasks include, among others, Cross-lingual STS (Semantic Text Similarity), Cross-lingual Retrieval as well as new forms of bitext mining, pointing towards the growth of relevance of diverse and multilingual resources. Despite these advancements, most large-scale Q\&A datasets used for supervised fine-tuning and evaluation of Language Models (LMs) are restricted to English and/or Wikipedia-derived, limiting their applicability in truly open-domain multilingual scenarios.

Previous work by Huber et al. addresses these limitations by introducing CCQA~\cite{Huber2022}, an open-domain question answering dataset derived from FAQ-style schema.org annotations. However, the dataset itself has not been publicly released, with only the data extraction code available. This limitation has hindered further research in leveraging FAQ-style QAs as a resource for fine-tuning or evaluation. To the best of our knowledge, no follow-up studies outside the original work have utilized CCQA, emphasizing the importance for publicly accessible datasets to support broa\-der advancements in the field. In this regard, we introduce WebFAQ, a large-scale openly available resource for ODQA and open-domain Q\&A retrieval, provided in a multitude of languages. WebFAQ encompasses 96 million QA pairs across 75 languages, including 47 million (49\%) non-English samples. The extracted QAs are further labeled with topic and question type.

% To address these limitations, we introduce WebFAQ, a large-scale openly available resource for ODQA and open-domain Q\&A retrieval, provided in a multitude of languages. Derived from FAQ-style schema.org annotations, WebFAQ encompasses 96 million Q\&A pairs across 75 languages, including 47 million (49\%) non-English samples. The extracted Q\&As are further labeled with topic and question type.
% Previous work by Huber et al~\cite{Huber2022} follows a similar methodology; however, their dataset CCQA has not been publicly released, with only the data extraction code available. This limitation has hindered further research in leveraging FAQ-style Q\&As as a resource for fine-tuning or evaluation. To the best of our knowledge, no follow-up studies outside the original work have utilized CCQA, emphasizing the importance for publicly accessible datasets to support broa\-der advancements in the field.

Building upon the WebFAQ base dataset, we introduce the following further key contributions:\\[5pt]
% \begin{itemize}
% \item
\textbf{Monolingual retrieval datasets}\\
We create 20 monolingual retrieval benchmarks with a total size of 11.2 million QA pairs (with 5.9 million non-English samples). To establish a clear notion of relevance within the retrieval datasets, refined filtering techniques have been applied, including near-dupli\-ca\-te detection and semantic consistency filtering for ques\-tion-ans\-wer pairs. The result is a well curated resource to be used in the context of multilingual dense retrieval.

To provide initial baselines on our proposed benchmarks, we measure the performance of BM25 and three state-of-the-art embedding models in a zero-shot setting. These evaluations assess mo\-del performance without any da\-ta\-set-specific adaptation. Separately, da\-ta\-set-specific fine-tuning is applied to an in-domain pretrained XLM-RoBERTa model using WebFAQ data.
The fine-tuned model achieves substantial performance gains, which generalize to other multilingual retrieval datasets.
% This highlights the potential of our dataset collection to enhance language models' understanding of relevance in open-domain Q\&A retrieval.
These improvements demonstrate that dense retrieval models benefit from exposure to WebFAQ data, leading to a concrete increase of model performance in open-do\-main Q\&A retrieval.\\[5pt]
% \item
\textbf{Bilingual datasets}\\
This is an entirely novel contribution with respect to the work of Huber et al. Using state-of-the-art bitext mining techniques and automated LLM-based translation evaluation, we have con\-struc\-ted 1k bilingual datasets containing a total of 1.5 million aligned QAs (with each of the 1001 language pairs comprising at least 100 QA pairs). This effort takes advantage of a unique opportunity: many FAQ pages exist in multiple languages, presenting the same questions and answers to users, but in translated form.
% This natural alignment allows for the extraction of parallel data on a large scale.
Most notably, the aligned text sequences of our final bitext corpora exhibit high translation quality, even when compared to human-curated bitext datasets, demonstrating the effectiveness of our approach for automated bitext generation.

\section{Related work}

% Web data commons

% QA datasets:

% web datasets: CCQA (also FAQ pages), Quora (Quora website), WebQuestions (Google Suggest API, small, answers are entities), ELI5 (Reddit), CoQA

% multilingual: TyDi QA (nur Wiki), MLQA (parallel corpus), XQUAD (translated), MKQA (translated)

% our dataset: extracted from heterogeneous websites, natural questions and answers, multilingual, more languages, no translations, can provide continuous updates to data

% Retrieval datasets:

% CLIRMatrix, XQUAD-R

% MS MARCO (Bing, web search queries), HotpotQA, QuoraRetrieval

% Parallel corpora, LASER, LaBSE, MuSR

% GEMBA

% - In the multilingual Q\&A space, datasets such as TyDi QA~\cite{Clark2020}, MLQA, XQuAD, and MKQA have contributed significantly to the development of cross-lingual question answering systems. TyDi QA is focused solely on Wikipedia-based questions, while MLQA, XQuAD, and MKQA include translated or parallel corpora that support cross-lingual tasks. However, these datasets are limited in size and scope compared to the multilingual and parallel corpora we propose.

\paragraph{QAs extracted from Common Crawl}
Our work builds upon the efforts of the Web Data Commons\footnote{\url{https://webdatacommons.org/}} (WDC) project, whose focus is the large-scale extraction of structured data from the Common Crawl\footnote{\url{https://commoncrawl.org/}} (CC) corpus. By systematically parsing and organizing sche\-ma.org annotations, embedded as JSON-LD, Microdata, RDF or Microformats, the WDC initiative provides the groundwork for various web data exploitations~\cite{Brinkmann2023}, such as the harvesting of FAQ pages. Accordingly, our work is indirectly inspired by CCQA \cite{Huber2022}, an open-domain question answering dataset from Meta AI, which also utilizes QA pairs extracted from Common Crawl. CCQA comprises approximately 55M unique QAs, including 24M English samples, gathered from 13 distinct web snapshots. In their paper, Huber et al. have demonstrated the effectiveness of CCQA for in-domain pre-training on tasks such as Closed-Book Question Answering (CBQA) and Passage Retrieval.

\paragraph{Q\&A datasets}
Beyond extracting QAs from Common Crawl, a variety of datasets from different sources have been developed to tackle distinct challenges in the context of question answering. Web\-Ques\-tions~\cite{Berant2013-WebQuestions}, built using the Google Suggest API, focuses on entity-based queries, while ComplexWebQuestions~\cite{Talmor2018} extends this sco\-pe by introducing broader and more complex questions requiring mul\-ti-step reasoning. ELI5~\cite{Fan2019}, sourced from Reddit, captures long-form explanatory Q\&A. CoQA~\cite{Reddy2019} shifts the focus to conversational Q\&A, modeling mul\-ti-turn context progression. Quora \cite{Sharma2019-arxiv} provides question pairs to evaluate semantic similarity rather than direct question answering. Additionally, the authors of PAQ (Probably Asked Questions)~\cite{Lewis2021} generated questions for selected Wi\-ki\-pe\-dia passages automatically, resulting in a large-scale collection of 65 million FAQ-style QA pairs.

Several large-scale Q\&A datasets have also played a central role in advancing information retrieval research. For instance, HotpotQA~\cite{Yang2018} introduces multi-hop reasoning, requiring models to retrieve and integrate information from multiple documents. Natural Questions (NQ)~\cite{Kwiatkowski2019} provides real user queries from Google Search along with corresponding Wikipedia passages, making it a widely used resource for training ODQA systems and Q\&A retrievers. Similarly, MS MARCO~\cite{Bajaj2016-arxiv}, a large-scale dataset derived from real Bing search queries, has become a standard benchmark for passage ranking and retrieval tasks.

% Another large-scale Q\&A datasets is PAQ, short for Probably Asked Questions, which comprises 65M automatically generated Q\&A pairs~\cite{Lewis2021}. The authors highlight the utility of directly leveraging Q\&A pairs for CBQA models and Q\&A retrievers, emphasizing their speed, interpretability, and update efficiency compared to retrieve-and-read systems. However, PAQ focuses on automatically generating questions to selected answer passage mined from Wikipedia, which, while valuable, differ fundamentally from the diverse, naturally occurring Q\&A pairs from FAQ pages annotated with schema.org.

Existing multlingual Q\&A datasets in the context of information retrieval include Mr. TyDi~\cite{Zhang2021}, MIRACL~\cite{Zhang2023} and MLDR~\cite{Chen2024-M3}. Mr. TyDi is a human-labeled retrieval dataset built on top of TyDi QA, which was sourced from Wikipedia and covers 11 typologically diverse languages. The limitations of Mr. Tydi, such as the methodology for annotating positive passages, were later addressed in MIRACL. These two datasets as well as the recent MLDR (Multilingual Long-Document Retrieval), with a focus on more lengthy sample texts, were specifically crafted to facilitate the training and evaluation of multilingual retrieval systems.

\paragraph{Cross-lingual datasets}
Significant IR datasets with queries and documents aligned across languages include mMARCO~\cite{Bonifacio2021-arxiv}, the multilingual version of the MS MARCO passage retrieval data\-set with the original English texts translated in 13 further langua\-ges. Further cross-lingual datasets are, e.g., MKQA~\cite{Longpre2021}, consisting of 10k QA pairs selected from NQ and translated from English into 25 additional languages, XQuAD~\cite{Artetxe2020-XQuAD}, the cross-lingual adoption of SQuAD, and CLIRMatrix~\cite{Sun2020}, currently the largest dataset in the context of Cross-Lingual IR (CLIR). CLIRMatrix includes 49M unique queries and 34 billion relevance labels across 139 languages with a massive bilingual corpus of 19,182 language combinations.
% Additionally, CLIRMatrix contains a jointly aligned dataset of que\-ries and documents across eight languages. However, the queries in CLIRMatrix are derived from Wikipedia article titles rather than natural language questions, limiting its applicability for natural Q\&A tasks.

% CLIR is one of many subtasks included in the Massively Multilingual Text Embedding Benchmark (MMTEB)~\cite{Anonymous2024-MMTEB}, which expands on the established MTEB benchmark~\cite{Muennighoff2022-arxiv}. MTEB evaluates multilingual models on diverse tasks, while MMTEB focuses on cross-lingual evaluation, including CLIR and parallel corpora mining. Parallel corpora are essential resources for machine translation, multilingual question answering, and cross-lingual retrieval tasks. By aligning text at the sentence or document level across languages, parallel corpora enable accurate and consistent model training and evaluation in multilingual settings.

\paragraph{Bitext mining}
The state-of-the-art approach for creating cross-lingual data\-sets is -- beyond translation -- automated sentence a\-lignment via similarity search over text embeddings. Notable methods include LASER (Language-Agnostic SEntence Representations) \cite{Artetxe2019-LASER} and its successor LaBSE (Language-Agnostic BERT Sentence Em\-bed\-dings) \cite{Feng2022}.
LASER emplys a sequence-to-sequence architecture for encoding, while LaBSE utilizes a Trans\-former-based model architecture.
These advancements in multilingual sentence embeddings contribute to reduced error rates in cross-lingual similarity search and thus enable efficient bitext mining.
Beyond bitext mining, recent research has explored the application of Large Language Models (LLMs) for automated translation evaluation. For example, Kocmi et al. introduce GEMBA \cite{Kocmi2023}, a GPT-based metric for translation evaluation, and demonstrate that LLMs can assess translation quality on par with human evaluators. These findings suggest that LLMs can play a critical role in both generating and validating cross-lingual datasets.

Notable datasets in the field of bitext mining include WMT 2019 \cite{wmt19translate}, a massive dataset of 124M bitext pairs spanning nine language combinations, introduced as part of the reocurring translation task at the Conference on Machine Translation (WMT). Another resource for bitext pairs is Tatoeba\footnote{\url{https://tatoeba.org/}}, a community-driven collection of sentences and their translations provided in a multitude of languages, parts of which are easily accessible through HuggingFace\footnote{\url{https://huggingface.co/datasets/Helsinki-NLP/tatoeba}}. Additionally, the BUCC 2018\footnote{\url{https://comparable.limsi.fr/bucc2018/bucc2018-task.html}} dataset, originating from the 11th Workshop on Building and Using Comparable Corpora (BUCC2018), contains 35k bitext pairs in four language combinations.

% Our work builds on these foundations by addressing gaps in existing datasets, particularly in the context of natural Q\&A pairs and parallel corpora. By leveraging schema.org annotations and an efficient three-step pipeline, we aim to provide a large-scale, multilingual, and high-quality dataset that facilitates research and development across multiple domains.

\section{Data Collection and Filtering}

This section describes the methodology used to develop the WebFAQ Q\&A dataset, including data collection, language detection, and topic and question type classification. Additionally, Section~\ref{refined_filtering} outlines the refined filtering techniques employed to transform the raw QA corpus into a high-quality retrieval dataset with well-defined relevance relationships between queries and docu\-ments.

\begin{figure*}[t]
\centering
\begin{minipage}[c]{0.25\textwidth}  % Adjust width as needed
\centering
\includegraphics[width=\linewidth]{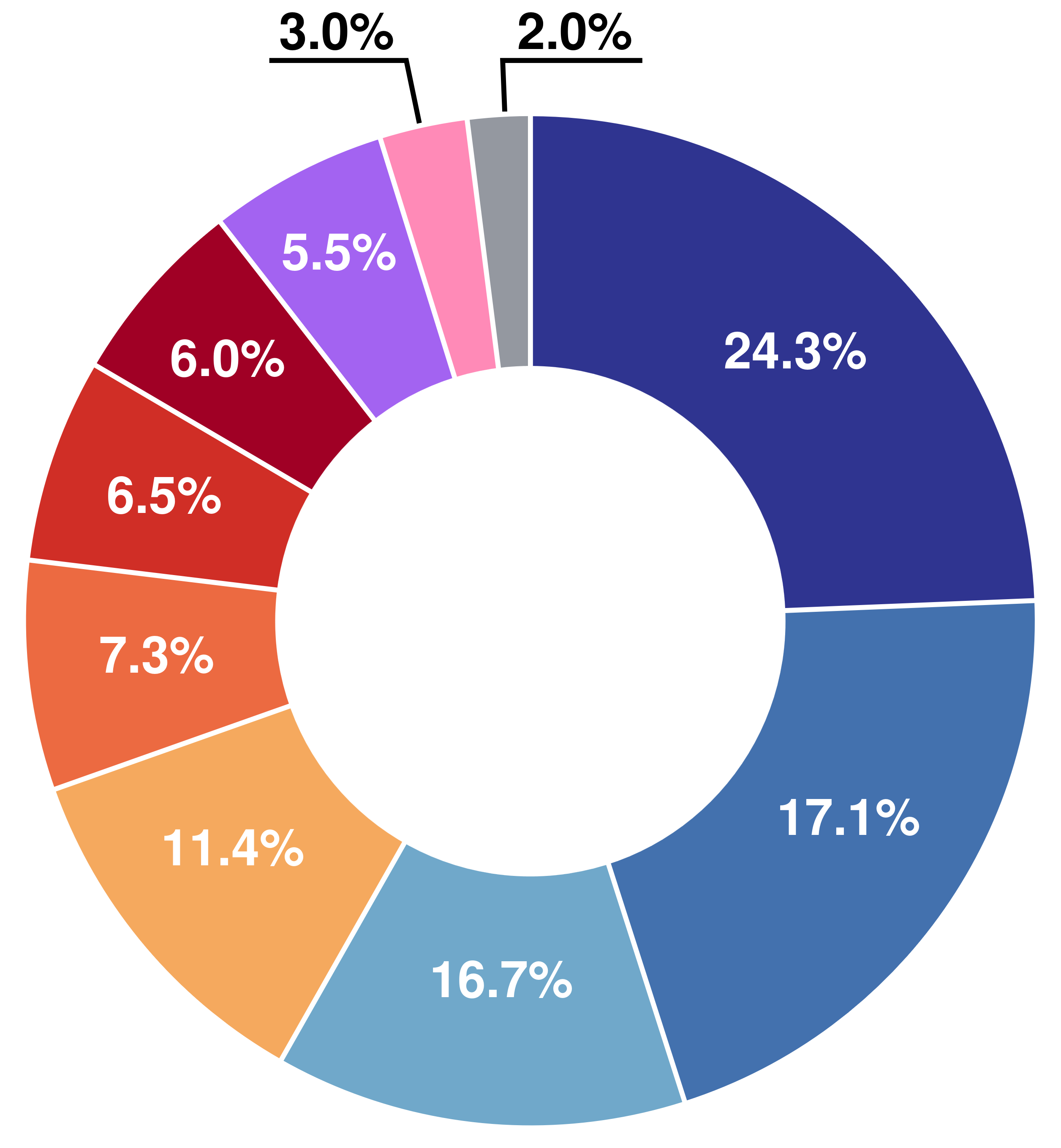}
\end{minipage}
\hfill
\begin{minipage}[c]{0.68\textwidth}  % Adjust width as needed
\centering
\fontsize{8pt}{12pt}\selectfont
\sffamily
\definecolor{color1}{HTML}{2F3490}
\definecolor{color2}{HTML}{4371AE}
\definecolor{color3}{HTML}{70A8CA}
\definecolor{color4}{HTML}{F5A95E}
\definecolor{color5}{HTML}{EC6A41}
\definecolor{color6}{HTML}{D02E26}
\definecolor{color7}{HTML}{A00025}
\definecolor{color8}{HTML}{A363F1}
\definecolor{color9}{HTML}{FF8AB7}
\definecolor{color10}{HTML}{9498A0}
\begin{tabular}{lll}
\cellcolor{color1} & \textbf{What} is a collision waiver? & Kaj je opustitev od trka? (slv) \\
\cellcolor{color2} & \textbf{How} is it like working at Coveo? & Comment c’est travailler chez Coveo? (fra) \\
\cellcolor{color3} & \textbf{Is, are, do, does} & Do I need to call ahead? (eng) \\
\cellcolor{color4} & \textbf{Which} sports are popular in Norway? & Hvilke idretter er populære i Norge? (nor) \\
\cellcolor{color5} & \textbf{Can, will, may, shall} & Can you guarantee ADA compliance? (eng) \\ % \textbf{Can, could, will, would, may, might, shall, should}
\cellcolor{color6} & \textbf{[No Question Word]} Fire Extinguishers & \begin{CJK}{UTF8}{min}消火器\end{CJK} (jpn) \\
\cellcolor{color7} & \textbf{When} will my test results come out? & Kapan hasil tes saya keluar? (msa) \\
\cellcolor{color8} & \textbf{Where} are the In-person Voting Points? & On estan els Punts de Votació Presencial? (cat) \\
\cellcolor{color9} & \textbf{Why} does Jekkle promo code not work? & \begin{CJK}{UTF8}{bsmi}為什麼Jekkle促銷碼無效?\end{CJK} (zho) \\
\cellcolor{color10} & \textbf{Who} is WordPress suitable for? & Für wen eignet sich WordPress? (deu) \\
\end{tabular}
\end{minipage}
% \vspace{-1.2cm}
\caption{Distribution of Question types with examples}
\Description{}
\label{fig:question_type}
\end{figure*}

\subsection{Data Source}
The raw data used to build WebFAQ originates from three Common Crawl snapshots, specifically the October dumps from 2022 to 2024. These web dumps were processed by the WDC initiative, which extracted structured schema.org annotations, including those marked with the \texttt{FAQPage} schema type. The extracted structured data is categorized into schema.org-specific subsets and made publicly available to support downstream research applications such as ours.

As noted by Huber et al.~\cite{Huber2022}, the use of schema.org metadata requires additional effort from website creators, implying that the annotated QA pairs are intended for public use and are therefore more likely to be relevant, well-formed, and informative. Despite the inherent noisiness of web data, their findings confirm that the vast majority of web-mined QA pairs are sensible and answerable. This strongly supports their use as resource for large-scale dataset creation.

\subsection{Processing}
The structured FAQ data is first parsed to extract question-answer pairs while removing boilerplate text, quotation marks, and emojis. The extracted texts are then subjected to basic deduplication and filtering of incorrectly formatted samples.
% This initial cleaning step ensures that only Q\&A content, which is intended as such, is retained.

\subsubsection{Language detection}
To classify the language of each QA pair, \texttt{fastText}~\cite{Joulin2017, Joulin2016-arxiv} is applied to the concatenated question-answer texts. In summary, the entire corpus comprises 75 languages with at least 1,000 samples each, while 49 languages appear in FAQ pages from at least 100 distinct websites\footnote{Websites are defined by their \textit{origin}, which includes scheme, host and optionally port.} (see Table~\ref{tab:languages}).
% 36 and 1,000

\begin{table}[h]
  \caption{Distribution of languages}
  \label{tab:languages}
  \vspace{-0.2cm}
  \begin{tabular}{p{2.5cm}rp{0.1cm}p{2.5cm}r}
  \toprule
  \textbf{Language} & \multicolumn{1}{c}{\textbf{\%}} & & \textbf{Language} & \multicolumn{1}{c}{\textbf{\%}} \\
  \midrule
  eng (English) & 51.2 & & ita (Italian) & 2.7 \\
  deu (German) & 6.9 & & jpn (Japanese) & 2.6 \\
  spa (Spanish) & 6.0 & & pol (Polish) & 1.7 \\
  fra (French) & 4.8 & & por (Portuguese) & 1.7 \\
  rus (Russian) & 3.8 & & tur (Turkish) & 1.5 \\
  nld (Dutch) & 2.8 & & Other & 13.0 \\
  \bottomrule
  \end{tabular}
\end{table}

\subsubsection{Topic and Question Type Classification}
To further analyze and categorize the collected FAQ data, we fine-tune \texttt{XLM-RoBERTa\-base}, a multilingual Transformer model, for two text classification tasks: (1) Topic Classification, (2) Question Type Classification. The resulting models are used to label the extracted QA pairs.

The respective training datasets of both classification tasks are composed of a sampled subset of QAs, which were automatically annotated using Open\-AI's \texttt{GPT-4o-mini}. To ensure sufficient training and validation data for each language, only those 49 languages were considered for labeling with at least 100 distinct websites contributing QA pairs. Additionally, to maximize diversity, the training sets include at most 1,000 QA pairs per language with no more than one sample per website, resulting in a final corpus of 37,383 samples.

\paragraph{Topic Classification}
The schema of tasks, provided to the classifier, is inspired by Curlie (formerly the Open Website Directory)\footnote{\url{https://curlie.org}}. However, because Curlie’s top-level categories focus more on commercial services and general information, we adapted its structure to better reflect the characteristics of FAQ-style question answering. Table~\ref{tab:topics} presents the final set of topics and their distribution across 96 million labeled QA pairs.

The dataset is split into 80\% training, 10\% validation, and 10\% test sets. After two epochs of fine-tuning, the F1-score on the validation set is 81.29\%.

\begin{table}[h]
  \caption{Distribution of topics}
  \label{tab:topics}
  \vspace{-0.2cm}
  \begin{tabular}{clr}
  \toprule
  \multicolumn{2}{l}{\textbf{Topic}} & \multicolumn{1}{c}{\textbf{\%}} \\
  \midrule
  \faPlane & Traveling and Hospitality & 34.1 \\
  \faShoppingCart & Products and Commercial Services & 19.8 \\
  \faHeartbeat & Healthcare Services, Wellness, and Lifestyle & 13.0 \\
  \faMusic & Entertainment, Recreation, and Leisure & 9.7 \\
  \faGraduationCap & Employment, Education, and Training & 9.5 \\
  \faBank & Banking, Financial Services, and Insurance & 6.0 \\
  \faGavel & Legal Services, Regulations, and Government & 4.0 \\
  \faInfoCircle & General Information and Other & 3.9 \\
  \bottomrule
  \end{tabular}
\end{table}

\paragraph{Question Type Classification}
For question type classification, we adopt a schema similar to previous studies~\cite{Fan2019, Yang2018}.
While these works rely on string matching to determine question types, this approach is impractical in a vastly multilingual setting. Instead, we use XLM-RoBERTa to correctly map non-English questions to their English-equivalent question words. For example, the Slovenian question ``Kaj je opustitev od trka?'' is classified under ``What'', without any custom definition of ``Kaj'' (\textit{What}) as keyword to be matched, but merely relying on the capabilities of the pretrained model to map correctly across languages.

Figure~\ref{fig:question_type} shows the distribution of question types among 96 million labeled QA pairs. Again, the dataset is split into 80\% training, 10\% validation, and 10\% test, achieving an F1-score of 78.56\% after three epochs of fine-tuning.

\subsection{Refined Filtering} \label{refined_filtering}
The initial processing pipeline produces a large-scale but relatively unclean dataset, containing ambiguous cases such as: (1) duplicate questions with different answers, (2) near-duplicate questions, and (3) generic QA pairs, describing those question or answer samples that require the context of their source webpage for meaningful interpretation.
% \begin{enumerate}
% \item Duplicate questions with different answers
% \item Near-duplicate questions
% \item Generic QA pairs – Cases requiring webpage context for meaningful interpretation.
% \end{enumerate}
Additionally, some answers may be factually incorrect, even if they remain topically relevant to the question. Note that ensuring factual accuracy is however beyond the scope of this paper.

\paragraph{Q\&A and Retrieval datasets}
We publish the full Q\&A dataset as an open resource\footnote{\url{https://huggingface.co/datasets/PaDaS-Lab/webfaq}}, allowing for custom filtering based on user needs. However, we find that datasets in the context of information retrieval have a more constrained notion of quality, as retrieval datasets must ensure that the concept of \textit{relevance} between query and corresponding documents is well reflected and not blur\-red through insufficiencies during data preparation. This is particularly true for datasets like WebFAQ, which rely on sparse relevance judgments that can be obscured by near-duplicates and similar artifacts.

\paragraph{Filtering}
We therefore propose three filtering techniques to eliminate the aforementioned ambiguities. The primary objective is to exclude QA pairs where a direct and unambiguous assignment between a question and its answer is not possible, coming at the cost of removing genuine QAs. This filtering involves:
\begin{enumerate}
\item Question-based Deduplication
\item Near-Duplicate Detection via Semantic Similarity Search
\item Question-Answer Semantic Consistency Filtering
\end{enumerate}

Step (1) is rather straightforward, eliminating those duplicate questions with different answers, whereas steps (2) and (3) are more intricate and require the computation of text embeddings for both questions and answers. For this cause, we use the multilingual embedding model Jina (v3), as it performs well on Semantic Textual Similarity (STS) as well as retrieval tasks across multiple languages~\cite{Sturua2024-arxiv} and ranges in the top ranks of common leaderboards.
% \footnote{\url{https://huggingface.co/jinaai/jina-embeddings-v3}}

\paragraph{Near-duplicate detection}
For (2), we establish a cosine similarity threshold $\alpha$ among questions of same website origin. Figure~\ref{fig:filtering_examples_1} presents examplary filtered near-duplicate questions with cosine similarity $\text{sim}_{Q_1, Q_2} > \alpha$.
% For instance, while the questions ``How to convert from JPEG to PDF?'' and ``How to convert from DOCX to TXT?'' are subjectively distinct, dense retrieval models often struggle to differentiate them.
% Similarly, question pairs that are merely reworded versions of each other, such as ``Does Chrome have a free VPN?'' and ``Is there a free VPN for Chrome?'', are also classified as near-duplicates.
% In both cases, establishing a \textit{unique} relationship between one question and its corresponding answer is challenging, leading to the elimination of these QA pairs.
For instance, question pairs that are merely reworded versions of each other, such as ``Does Chrome have a free VPN?'' and ``Is there a free VPN for Chrome?'', are classified as near-duplicates. Since a \textit{unique} relationship between one question and its corresponding answer cannot be established, these QA pairs are eliminated.

\paragraph{Filtering based on semantic consistency between QAs}
For (3), we define a minimum cosine similarity threshold $\beta$ between questions and their corresponding answers. QAs falling below this threshold are removed ($\text{sim}_{Q, A} < \beta$). For instance, Figure~\ref{fig:filtering_examples_2} illustrates a case where the question lacks a specific entity (``Can I test \textit{it} before purchasing?''). Without the context of the web page or the answer, it is unclear that ``\textit{it}'' refers to an SMS service. As a result, the question by itself remains too generic, making it unrealistic to expect any state-of-the-art retriever to accurately determine its relevance to the corresponding answer.

\begin{figure}[h]
\fontsize{9pt}{11pt}\selectfont

\begin{subfigure}{\linewidth}
    \centering
    \sffamily
    \begin{tabular}[c]{@{}cp{5.5cm}c@{}}
    \toprule
    %  & $\text{Q}_1$: How to convert from JPEG to PDF? &  \\
    %  & $\text{Q}_2$: How to convert from DOCX to TXT? &  \\
    % \midrule
     & $\text{Q}_1$: Does Chrome have a free VPN? &  \\
     & $\text{Q}_2$: Is there a free VPN for Chrome? &  \\
    \bottomrule
    \end{tabular}
    \caption{Near-duplicate detection ($\alpha = 0.7$)}
    \label{fig:filtering_examples_1}
\end{subfigure}

\vspace{0.5cm}

\begin{subfigure}{\linewidth}
    \centering
    \sffamily
    \begin{tabular}[c]{@{}cp{5.5cm}c@{}}
    \toprule
     & Q: Can I test \textit{it} before purchasing? &  \\
     & A: Yes, you send 10 SMS campaigns to &  \\
     & \:\:\:\: test our SMS service for free. &  \\
    \bottomrule
    \end{tabular}
    \caption{QA Semantic Consistency Filtering ($\beta = 0.5$)}
    \label{fig:filtering_examples_2}
\end{subfigure}

% \vspace{-0.2cm}
\caption{Examples of filtered questions/QA pairs}
\label{tab:filtering_examples}
\end{figure}

It is important to note that the goal of the proposed filtering steps is not to remove hard negatives, which can enhance a retrieval dataset -- though this might be a side effect of the filtering process. Instead, the primary objective is to exclude QA pairs where state-of-the-art dense retrievers cannot reliably distinguish answers as either positive or negative.
% Any fees or regulations that you usually have to pay when you use your Visa card online will still apply, as per your agreement with the company. However, the casinos that we recommend will never charge you extra simply for using your credit card.

The parameters $\alpha = 0.7$ and $\beta = 0.5$ were determined through manual inspection of 2,000 samples from the English subset, prioritizing dataset quality over sheer size.
Applying those filtering steps to our collected QAs led to the creation of 20 large-scale retrieval corpora\footnote{\url{https://huggingface.co/datasets/PaDaS-Lab/webfaq-retrieval}}, derived from the 20 largest language subsets, with a total size of 11.2 million QA pairs (5.9 million non-English).
For the time being, we restricted the number to 20 subsets, such that each individual corpus contains at least around 80k distinct samples, ensuring a robust and diverse foundation for retrieval tasks.

% as these have a minimum size of at least 79k distinct samples.
% each of the 20 containing at least 100,000 distinct samples.

\section{Evaluation} \label{evaluation}

This section demonstrates the effectiveness of the WebFAQ retrie\-val datasets through experiments in an open-retrieval setting.
First, we establish retrieval performance baselines by evaluating BM25 alongside three state-of-the-art embedding models and compare their results on WebFAQ to those obtained on other multilingual datasets. Second, we fine-tune an in-domain pre\-trained language model using WebFAQ data.
We hypothesize that this domain-spe\-ci\-fic fine-tuning step enhances the model’s understanding of relevance in retrieval tasks by leveraging supervised learning from WebFAQ’s QA pairs. As a result, substantial performance improvements across languages are expected, with potential benefits extending to other datasets in a true zero-shot setting.
% In order to consider our resource useful, we hypothesize that through this step of data-specific fine tuning we can achieve substantial performance gains across languages. These improvements may also transfer to other datasets evaluated in a truly zero-shot setting.

All models are evaluated on the WebFAQ retrieval collection alongside two common multilingual retrieval datasets, namely Mr. TyDi \cite{Zhang2021} and MIRACL \cite{Zhang2023}. Both originate from Wikipedia-based sources and are widely used for training and evaluating multilingual retrievers. For MIRACL, we evaluate on the smaller hard negatives subset, which is part of the recent MMTEB leaderboard. % to provide a more robust comparison.
Table~\ref{tab:results} outlines retrieval performances on six languages -- the intersection set of languages covered by WebFAQ, Mr. TyDi and MIRACL. The full results are available in Appendix~\ref{appendix}.

% \begin{table}[b]
% \caption{Comparing retrieval performance of SoTA embedding models on 3 multilingual datasets using NDCG@10 (\%).}
% \label{tab:results1}
% \vspace{-0.2cm}
% \begin{tabular}{rcccccc}
%  & ara & eng & ind & jpn & kor & rus \\ \cline{2-7} 
%  &  &  &  &  &  &  \\[-9pt] 
%  & \multicolumn{6}{c}{WebFAQ} \\ \cline{2-7} 
% % \multicolumn{1}{r|}{BM25} &  &  &  &  &  &  \\
% mGTE (base) & 71.1 & 60.2 & 76.3 & 61.8 & 75.5 & 58.5 \\
% mE5 (large) & 80.0 & \textbf{67.1} & 82.1 & 73.2 & 83.4 & 68.4 \\
% Jina (v3) & \textbf{85.5} & 67.0 & \textbf{85.2} & \textbf{75.2} & \textbf{87.4} & \textbf{72.6} \\ \cline{2-7} 
%  &  &  &  &  &  &  \\[-9pt] 
%  & \multicolumn{6}{c}{MIRACL (Hard Negatives)} \\ \cline{2-7} 
% % \multicolumn{1}{r|}{BM25} &  &  &  &  &  &  \\
% mGTE (base) & 71.8 & \textbf{54.9} & \textbf{50.5} & \textbf{66.4} & 63.9 & 63.9 \\
% mE5 (large) & \textbf{75.6} & 46.7 & 50.4 & 66.1 & \textbf{65.8} & 63.5 \\
% Jina (v3) & 72.2 & 52.1 & 49.4 & 66.3 & 64.3 & \textbf{65.4} \\ \cline{2-7} 
%  &  &  &  &  &  &  \\[-9pt] 
%  & \multicolumn{6}{c}{Mr. Tydi} \\ \cline{2-7} 
% % \multicolumn{1}{r|}{BM25} &  &  &  &  &  &  \\
% mGTE (base) & 73.1 & \textbf{57.1} & 67.8 & 59.9 & 56.5 & 63.7 \\
% mE5 (large) & \textbf{76.5} & 52.5 & \textbf{71.3} & \textbf{61.9} & \textbf{59.5} & \textbf{65.4} \\
% Jina (v3) & 71.9 & 55.2 & 69.4 & 59.0 & 55.6 & 62.3 \\ \cline{2-7} 
% \end{tabular}
% \end{table}

\begin{table*}[h]
\caption{Comparing retrieval performance on 3 multilingual datasets using NDCG@10 in \%, including SotA embedding models and BM25 as baselines. \textbf{Base} and \textbf{FT} represent the pretrained XLM-RoBERTa model without/with fine-tuning on WebFAQ data. \textbf{Hybrid} combines BM25 and FT as described in Section~\ref{model_fine_tuning}. Bold font indicates top values w.r.t. the first 3 and last 4 rows.}
\label{tab:results}
\vspace{-0.2cm}
\addtolength{\tabcolsep}{-0.07em}
\begin{tabular}{rcccccccccccccccccccc}
 & \multicolumn{6}{c}{WebFAQ} &  & \multicolumn{6}{c}{MIRACL (Hard Negatives)} &  & \multicolumn{6}{c}{Mr. Tydi} \\ \cline{2-7} \cline{9-14} \cline{16-21} 
 &  &  &  &  &  &  &  &  &  &  &  &  &  &  &  &  &  &  &  &  \\[-10pt] 
 & ara & eng & ind & jpn & kor & rus &  & ara & eng & ind & jpn & kor & rus &  & ara & eng & ind & jpn & kor & rus \\ \midrule[1.5pt] % \cline{2-21} 
 &  &  &  &  &  &  &  &  &  &  &  &  &  &  &  &  &  &  &  &  \\[-10pt] 
% BM25 & 30.13 & 24.43 & 35.61 & 29.19 & 30.15 & 20.79 &  & 53.23 & 31.58 & 48.65 & 51.60 & 44.73 & 37.59 &  & 43.39 & 19.94 & 50.53 & 24.98 & 24.79 & 29.73 \\ 
mGTE & 71.1 & 60.2 & 76.3 & 61.8 & 75.5 & 58.5 &  & 71.8 & \textbf{54.9} & \textbf{50.5} & \textbf{66.4} & 63.9 & 63.9 &  & 73.1 & \textbf{57.1} & 67.8 & 59.9 & 56.5 & 63.7 \\
mE5 & 80.0 & \textbf{67.1} & 82.1 & 73.2 & 83.4 & 68.4 &  & \textbf{75.6} & 46.7 & 50.4 & 66.1 & \textbf{65.8} & 63.5 &  & \textbf{76.5} & 52.5 & \textbf{71.3} & \textbf{61.9} & \textbf{59.5} & \textbf{65.4} \\
Jina & \textbf{85.5} & 67.0 & \textbf{85.2} & \textbf{75.2} & \textbf{87.4} & \textbf{72.6} &  & 72.2 & 52.1 & 49.4 & 66.3 & 64.3 & \textbf{65.4} &  & 71.9 & 55.2 & 69.4 & 59.0 & 55.6 & 62.3 \\ \midrule[1.5pt]
BM25 & 30.1 & 24.4 & 35.6 & 29.2 & 30.2 & 20.8 &  & 53.2 & 31.6 & 51.6 & 44.7 & 37.6 & 29.8 &  & 43.4 & 19.9 & 50.5 & 25.0 & 24.8 & 29.7 \\ \midrule % \cline{2-7} \cline{9-14} \cline{16-21} 
Base & 61.0 & 49.7 & 74.6 & 56.5 & 73.2 & 50.1 &  & 36.1 & 30.6 & 28.1 & 27.6 & 39.3 & 29.6 &  & 30.5 & 18.1 & 33.2 & 16.0 & 27.7 & 22.5 \\
FT & \textbf{74.1} & \textbf{59.0} & \textbf{82.7} & \textbf{68.8} & \textbf{81.6} & \textbf{61.5} &  & 49.3 & 36.9 & 36.3 & 38.5 & 41.2 & \textbf{38.5} &  & 44.2 & 28.7 & 48.4 & 31.1 & \textbf{34.8} & 29.5 \\ \midrule
Hybrid & 32.4 & 28.6 & 37.4 & 32.0 & 31.8 & 24.4 &  & \textbf{61.7} & \textbf{50.8} & \textbf{57.8} & \textbf{51.3} & \textbf{43.2} & 35.3 &  & \textbf{54.4} & \textbf{31.7} & \textbf{58.5} & \textbf{37.8} & 30.4 & \textbf{36.0} \\ \bottomrule[1.5pt]
\end{tabular}

% \vspace{0.2cm}

% \begin{tabular}{rC{1.5cm}C{1.5cm}C{1.5cm}C{1.5cm}C{1.5cm}C{1.5cm}c}
%  & ArguAna & HotpotQA & MSMARCO & NQ & Quora & SCIDOCS & TRECCOVID \\ \cline{2-8} 
% BM25 & 59.3 & 33.8 & 49.0 & 38.2 & 44.1 & 21.5 & 61.2 \\
% Hybrid & \textbf{62.7} & 31.8 & \textbf{55.9} & \textbf{46.7} & 44.3 & \textbf{22.0} & \textbf{73.9} \\ \cline{2-8} 
% XLM-R (pre-trained) & 36.1 & 33.3 & 33.3 & 27.0 & 73.1 & 9.3 & 30.0 \\
% XLM-R (fine-tuned) & 45.5 & \textbf{37.2} & 28.5 & 29.0 & \textbf{83.2} & 11.3 & 52.7 \\
% \end{tabular}
\end{table*}

\subsection{Initial Baselines}

\paragraph{BM25}
To begin with, the reported results in Table~\ref{tab:results} include BM25, a strong traditional information retrieval method. Concre\-tely, we use the Pyserini implementation~\cite{Lin2021} built on the Lucene search library to generate a BM25 index per language and dataset. The default settings of Pyserini were not changed, employing Lu\-cene's default whitespace analyzer and setting hyperparameter values of $k_1=0.9$ and $b=0.4$.
% The retrieval results of BM25 in Table~\ref{tab:results} are interpreted in the following section.

\paragraph{SotA Embedding Models}
The baseline experiment further inclu\-des three state-of-the-art multilingual embedding models:
\begin{itemize}
\item \textit{mGTE}: GTE-Multilingual-Base~\cite{Zhang2024} (305M parameters)
\item \textit{mE5}: Multilingual-E5-Large-Instruct~\cite{Wang2024-arxiv} (560M parameters)
% \item BGE-M3~\cite{Chen2024-M3} (568M parameters)
\item \textit{Jina}: Jina (v3)~\cite{Sturua2024-arxiv} (572M parameters)
\end{itemize}

These models have demonstrated strong retrieval performance across various benchmarks and languages. They were selected as they had been -- at the point of model selection -- the three best ranked, public multilingual text embedding models on the MTEB leaderboard with less than 1B parameters.

\paragraph{Interpretation}
The findings in Table~\ref{tab:results} reveal inconsistent performance across datasets for the three aforementioned dense retrievers, while all of them perform relatively well compared to BM25. Jina, the largest model in terms of parameters, achieves the best results on WebFAQ but is outperformed on MIRACL and Mr. TyDi. This suggests that mGTE and mE5 likely benefit from prior exposure to the training splits of these datasets, indicating that their performance is influenced by the training data used during model development. According to the technical reports, mGTE and mE5 were trained using MIRACL and Mr. TyDi during supervised fine-tuning, whereas there is no explicit confirmation for Jina.

Beyond establishing baselines, the results further validate that WebFAQ's retrieval dataset collection meets two key requirements. First, it is discriminative, effectively differentiating retrieval performance across models based on NDCG@10. Second, model rankings remain relatively stable across languages, demonstrating the dataset’s robustness and consistency in multilingual evaluation.

\subsection{Model Fine-tuning} \label{model_fine_tuning}
To empirically study its efficacy, we utilize WebFAQ to fine-tune a pretrained multilingual Transformer model.
% The fine-tuning may demonstrate that a dense retrieval model benefits from exposure to WebFAQ's QAs during training, leading to an increase of retrieval performance across datasets.
Specifically, we again employ \texttt{XLM-RoBERTa-base} as foundation model and apply task-spe\-ci\-fic pretraining with MS MARCO~\cite{Bajaj2016-arxiv} data to create a base retrieval. While MS MARCO is a well-estab\-lished training dataset for information retrieval, its corpus consists solely of English texts. Thus, this initial pretraining phase does not directly incorporate multilingual data but serves as an extensive in-domain warm-up before multilingual fine-tuning.

During this pretraining stage, Margin MSE~\cite{Hofstätter2020-arxiv} is employed as a contrastive loss function, using hard negatives scored by a MiniLM cross-encoder, provided by the Sentence-Transformers framework. The official Sentence-Transformers training script\footnote{\url{https://sbert.net/examples/training/ms_marco/README.html}} is used for consistency and reproducibility. The model is trained for 30 epochs with 503k data samples and a batch size of 32.

After the task-specific pretraining on MS MARCO, we further fine-tune the model using the train splits of the WebFAQ retrieval datasets to better adapt it to the multilingual retrieval setting. The model undergoes an additional 20 epochs of training with 2,560,000 data samples and a batch size of 128, using Multiple Negatives Ranking Loss~\cite{Henderson2017-arxiv} with in-batch-negatives. The fine-tuning dataset is distributed relatively evenly across 20 languages, corresponding to the 20 retrieval corpora that we created. Each language contributes between 64,000 and 256,000 samples, depending on the size of the original train split. This ensures that the model is exposed to a diverse set of training samples.

\subsection{Results}
As shown in Table~\ref{tab:results}, fine-tuning with WebFAQ consistently improves retrieval performance. This effect is most evident in WebFAQ test splits, where the fine-tuned model achieves an average relative increase of 17\% in NDCG@10 across six languages. Additionally, the model demonstrates improvements in zero-shot retrieval tasks, where evaluation datasets were not encountered during task-specific pretraining or fine-tuning. The observed gains on Mr. TyDi and MIRACL (Hard Negatives) suggest that exposure to WebFAQ’s diverse multilingual QA pairs enhances the model’s ability to judge relevance more effectively. These findings confirm the hypothesis outlined in the first paragraph of Section~\ref{evaluation}.

\paragraph{BM25}
When comparing the fine-tuned model with BM25, we find that our model outperforms the sparse retrieval approach on WebFAQ. However, on Mr. TyDi and MIRACL, the performance gap between BM25 and the fine-tuned dense retriever narrows. This outcome is expected, as these datasets are evaluated in a zero-shot setting. As noted by Zhang et al.~\cite{Zhang2021}, dense retrievers typically struggle in zero-shot retrieval scenarios due to their reliance on in-domain training for learning effective representations.

\paragraph{Hybrid}
%TODO: Describe the exact method used to combine dense and sparse retrieval results.
% We experiment with a hybrid retrieval approach, which combines sparse (BM25) and dense retrieval (Fine-tuned XLM-Ro\-ber\-ta) methods. To this end, we first perform the regular retrieval step for a query once with the sparse and dense model, respectively, retrieving the top 1000 documents. We then merge the retrieved documents by calculating a combined similarity score for query $q$ and document $D_i$ as proposed by Ma et al.~\cite{Ma2021-arxiv}:
Lastly, we investigate whether our fine-tuned model retains useful relevance signals, even in cases where BM25 outperforms the dense retriever. To explore this, we employ a hybrid retrieval approach that integrates sparse (BM25) and dense (WebFAQ-trained XLM-RoBERTa) retrieval methods. The hybrid mo\-del first performs the regular retrieval step for both the sparse and dense model, retrieving the top 1000 documents, respectively. The retrieved sets are then merged by computing a combined similarity score for query $q$ and document $D_i$, following Ma et al.~\cite{Ma2021-arxiv}:
\begin{equation}
    \lambda sim(q, D_i) + BM25(q, D_i)
\end{equation}
where $sim(q, D_i)$ represents the cosine similarity between $q$ and $D_i$ using XLM-RoBERTa, and BM25 provides the corresponding sparse retrieval score. Our implementation follows the original paper in setting $\lambda=1.1$. If a document appears in only one retrieval set, its missing score is assigned as zero. Our findings indicate that the fine-tuned model continues to capture valuable relevance signals, enhancing retrieval performance beyond the BM25 baseline in several cases, even when BM25 alone outperforms the dense retriever.

\section{Bilingual datasets}

% - Generating candidates using embeddings should work because multilingual embedding models are trained with the objective of cross-lingual alignment. That means they can hopefully align passages of different languages in roughly the same region of the embedding space. Even though sentences of different languages might be located in different manifolds (or language-specific subspaces), their vector embeddings should still lay close to each other.
% - After generating candidates using a threshold, we evaluate the candidates with GPT-4o.

% Website is the smallest entity ...

% "https://scubanana.es/faq/", "https://scubanana.es/de/haeufig-gestellte-fragen/"

In addition to constructing a collection of monolingual retrieval corpora, we further utilize the collected QAs to create QA-alig\-ned bitext datasets. Our approach exploits the fact that many websites provide FAQ pages in multiple languages, where the questions and answers are often literal translations of each other, since they aim to offer consistent services disregarding the users' provenance.

\subsection{Methodology}
The extraction of aligned text pairs employs state-of-the-art bitext mining techniques and utilizes cross-lingual similarity search based on sentence embeddings. Specifically, we use LaBSE (Lang\-uage-Ag\-nos\-tic BERT Sentence Embeddings)~\cite{Feng2022} for bitext candidate generation. LaBSE generates vector representations of text sequences and enables cross-lingual similarity computation via cosine similarity scores. On top of this, the candidates' translation quality is assessed using GEMBA, a GPT-based metric introduced by Kocmi et al~\cite{Kocmi2023}. Within our method, GEMBA is used to automatically annotate bitext candidates with a binary label (text pair is \textit{accepted} / \textit{rejected}) and find a reasonable cutoff value for cosine similarity between cross-lingual vector embeddings of candidate pairs. During alignment, the concatenation of a question and its corresponding answer is treated as a single unit, and bitexts are thus question-answer pairs.

The bitext mining process follows these key steps:
\begin{enumerate}
\item \textbf{Elimination of near-duplicate QAs.}\\
The chosen approach of bitext mining includes cross-lingual similarity search, which is prone to similar text sequences within one language. Due to the large number of cross-lingual matchings of near-duplicates, the volume of generated bitext candidates had grown to an extremely large extent. To overcome this limitation, we applied the same elimination of near-duplicate QAs as described in Section~\ref{refined_filtering}.

\item \textbf{Computation of LaBSE embeddings} for all text sequences (question + answer) of the 75 monolingual Q\&A datasets.
% assume a collection of monolingual Q\&A corpora, where each corpus is accompanied by its LaBSE vector representations.

\item \textbf{For each language pair: Computation of cosine similarity values between cross-lingual text pairs.}\\
% through cross-lingual similarity search.\\
The matching of text pairs is conveyed across languages and restricted to website level. A text of language $L_1$ is thus only matched with those texts of language $L_2$ originating from the same website. This covers also those multilingual FAQs exposed under different language-specific URL paths\footnote{Example of English FAQ page with German counterpart under a different URL path:\\\url{https://scubanana.es/faq/} and \url{https://scubanana.es/de/haeufig-gestellte-fragen/}}.
Furthermore, to limit the volume of candidates to a reasonable size, all text pairs with cosine similarity below a tolerance threshold of $s=0.80$ are discarded.

\item \textbf{For each language pair: Annotation of a sampled subset of text pairs as \textit{accepted} or \textit{rejected} translations.}\\
The remaining bitext candidates are randomly sampled with a probability of 0.1\% and, if selected, automatically annotated with regards to their translation quality.
This task is performed using \texttt{GPT-4o-mini} evaluating according to the GEMBA metric, which assigns a direct assessment (DA) score between 0 and 100. As observed by Kocmi et al, LLM-based DA scores are not uniformly distributed, as the model exhibits preference biases in assigning specific values.

Through manual inspection, we find that scores between 80 and 84 are rarely used (1\%), whereas 85 (16\%), 90 (18\%) and 95 (51\%) are frequently assigned by \texttt{GPT-4o-mini} to indicate high-confidence translations. Based on this observation, we classify bitexts with scores $\ge 85$ as correct translations, while those below 85 are considered incorrect.

\item \textbf{Definition of cosine similarity threshold.}\\
The LLM-generated annotations are used to define a final threshold with respect to a desired level of precision. For the concrete case of our dataset, we find that a threshold of $s=0.90$ provides a reasonable balance between quality and recall, yielding 95\% correct translations according to the LLM-based assessment.
\end{enumerate}

This method yields a total amount of 1.5 million aligned QA pairs across 1176 language pairs. For publishing our collection of bilingual datasets\footnote{\url{https://huggingface.co/datasets/PaDaS-Lab/webfaq-bitexts}}, all language pairs featuring less than 100 samples were omitted, leaving 1,487,328 aligned texts across 1001 combinations. The most frequent language pairs are German-English (37,348 text pairs), followed by English-French (37,208) and English-Spanish (35,446).

% $\sigma_{pooled}^{2} = \frac{\sum{(|S_{l_1, l_2}| - 1) \times \sigma_{i}^{2}}}{\sum{|S_{l_1, l_2}| - 1}}$
\begin{table}[h]
\caption{Comparison of average translation quality scores for different bitext corpora. Pooled variance captures the variance of scores across language pairs.}
\label{tab:bitext_datasets}
\begin{tabular}{ccccc}
\toprule
 & Size & \#Pairs & \begin{tabular}[c]{@{}c@{}}Translation quality\\Avg. score ($\pm$ std)\end{tabular} & \begin{tabular}[c]{@{}c@{}}Pooled\\variance\end{tabular} \\ \midrule
WMT 2019 & 124M & 9 & 85.1 $\pm$ 13.4 & 173.1 \\
BUCC 2018 & 35.0k & 4 & 85.9 $\pm$ 12.6 & 158.7 \\
Tatoeba & 88.9k & 113 & 77.3 $\pm$ 30.2 & 517.4 \\ \midrule
mMARCO & 8.8M & 182 & 86.6 $\pm$ 11.2 & 123.9 \\ \midrule
\begin{tabular}[c]{@{}c@{}}WebFAQ\\(Bitexts)\end{tabular} & 1.5M & 1001 & \textbf{91.0 $\pm$ 8.6} & \textbf{73.1} \\ \bottomrule
\end{tabular}
\end{table}

\subsection{Quality Evaluation}
To validate the presented bitext mining method, the resulting data\-set is compared to existing parallel corpora in terms of average translation quality. Therefore, a random sample of 20,000 bitext pairs is selected from each corpus and evaluated based on the GEM\-BA metric, assigning scores on a 0 to 100 scale. The final average scores per bitext dataset are listed in Table~\ref{tab:bitext_datasets}.

Among the evaluated bitext datasets, WMT 2019 stands out as the largest, containing 124 million bitext pairs across nine language pairs. The corpus extracted from WebFAQ is two orders of magnitude smaller, but yet surpasses both BUCC 2018 and Tatoeba in terms of dataset size.
Regarding translation quality, our dataset achieves the highest average GEMBA score and the lowest standard deviation. This suggests that the cosine similarity thresholding for cross-lingual sentence embeddings, described in the methodology, effectively produces a large-scale bitext corpus that remains comparable to existing resources in terms of translation accuracy. The six-point gap in scores compared to WMT 2019 and BUCC 2018 supports this claim. We believe that this holds true despite potential biases being introduced by using the same approach for LLM-based translation assessment for both dataset generation and evaluation.
Additionally, our findings are corroborated by the comparison with mMARCO, an automatically translated retrieval dataset, which was created witha focus on translation quality, yet exhibits significantly lower GEMBA scores than our dataset.
Overall, these findings confirm the effectiveness of our approach combining cross-lingual sentence embeddings with automated quality assessment in mining cross-lingual QA pairs.

\section{Discussion}

The following section discusses several aspects that should be considered when using WebFAQ for research and evaluation. The first aspect is the prevalence of \textit{sparse relevance judgments} in the retrieval datasets (Section~\ref{refined_filtering}). Generally speaking, this prevalence entails that no claim can be made about the exhaustiveness of the dataset's relevance annotations. Even though this is a widely accepted practice in IR research (see e.g., MS MARCO~\cite{Bajaj2016-arxiv} and Mr. Tydi~\cite{Zhang2021}), one should keep in mind that the ``qrels'' of our dataset provide only one good answer to a question, and leave out many more answers that are also potentially relevant. This is particularly true for rather generic QAs, such as ``What is the average time for my package to be delivered?'', which is asked and answered by different delivery service FAQs in a similar fashion.
% , which do not account for generic questions that appear across multiple websites. Since our filtering process prioritizes unique QA pairs, generic questions -- such as those frequently asked about popular topics -- are not explicitly marked as relevant across different sources. As a result, retrieval models trained on this dataset may not fully capture the distribution of widely recurring queries, potentially limiting their ability to generalize in scenarios where such redundancy is meaningful.

Another discussion point arises with the the rigorous elimination of near-duplicate QA pairs, which primarily affects questions that revolve around entities. While we found this step helpful for removing many QAs one could call ``spam'' or low-quality, it may introduce an unintended bias by filtering out subtly different entity-based queries. For instance, the two questions ``What is CBD?'' and ``What is CBDa?'' are treated as near-duplicates and are thus removed, despite their subjective distinctions. This decision impacts analysis of retrieval models, as sparse methods (e.g., BM25) and dense models handle such minor lexical variations differently. The dataset, therefore, may not fully reflect scenarios where entity-level distinctions play a significant role in retrieval performance.

Finally, the quality of multilingual and cross-lingual datasets is inherently constrained by the accuracy of the language identification model. Through manual inspection, we observed systematic errors where the language detection system struggles with queries or answers containing long named entities from multiple languages. Additionally, in some edge cases, the question and answer are formulated in different languages, which the current identification method does not explicitly address, as it applies detection to the concatenated QA pair rather than handling them separately.
% These artifacts suggest that further refinements in language identification and multilingual data processing could improve dataset accuracy and retrieval performance across diverse linguistic contexts.

\section{Conclusion}

%Future work: Enriching the dataset with data from the Open Web Index
%These FAQPage annotations are mined using two sources of archived web content. The first one is a static data source, whereas the second one complements the first by enriching it with always new and fresh Question and Answer pairs that have been recently collected.
%Beyond that, we harness a second source of archived web content, namely the Open Web Index (OWI), which emerged from a recent European research initiative providing processed web content, formatted as index shards accompanied with Metadata files, in daily frequency. The OWI is collected using the Open Web Crawler, a continuously iterating crawling system that prioritizes thwe fetching of fresh web content by tracking the change frequency of the corresponding web pages.
%We furthermore aggregate the daily OWI index shards for one week and parse the collected Microdata and JSON-LD data in order to extract pairs of Question and Answers.

This paper introduces WebFAQ, a large-scale, multilingual resource for open-domain question answering and Q\&A retrieval. Derived from FAQ-style schema.org annotations, WebFAQ encompasses 96 million natural QA pairs across 75 languages. With our work, we offer publicly available data that surpasses previous efforts in scale and linguistic diversity, and promises to surpass them in terms of usability and practical impact.

To demonstrate the utility of WebFAQ, we curated 20 monolingual retrieval benchmarks, applying advanced filtering techniques to ensure an unblurred notion of relevance reflected in the datasets. Our empirical results show that fine-tuning a text embedding mo\-del on WebFAQ's QAs leads to significant retrieval performance improvements, with gains that generalize to other multilingual retrieval tasks. Additionally, we introduced a novel set of bilingual Q\&A datasets, constructed through state-of-the-art bitext mining and automated LLM-based translation evaluation. These bilingual corpora exhibit superior translation quality compared to existing datasets, which demonstrates the effectiveness of our automated approach for bitext generation.

With WebFAQ, we provide a publicly available, large-scale resource that facilitates the training and evaluation of multilingual retrieval models. We hope this work contributes to the diversification of evaluation tasks and training datasets and, in this regard, paves the way for broader advancements in open-domain multilingual Q\&A retrieval.

\begin{acks}
This work has received funding from the European Union's Horizon Europe research and innovation program under grant agreement No 101070014 (OpenWebSearch.EU, \url{https://doi.org/10.3030/101070014}).
\end{acks}

\bibliographystyle{ACM-Reference-Format}
\bibliography{sources}

\onecolumn

\appendix

\section{Appendix: Full Results} \label{appendix}
The below tables list the NDCG@10 in \% results across models for all 20 WebFAQ retrieval corpora, the 18 languages included in MIRCAL (Hard Negatives) and the 11 languages of Mr. Tydi. The top three rows show the performance of three state-of-the-art dense embedding models: GTE-Multilingual-Base (mGTE), Multilingual-E5-Large-Instruct (mE5) and Jina (v3). The bottom four rows of each table compare the results of BM25, a popular traditional sparse embedding model, which we use as a baseline, to our pretrained XML-RoBERTa models. Here, \textbf{Base} refers to the model with in-domain pretraining on MS MARCO, while \textbf{FT} refers to the model that was additionally fine-tuned on WebFAQ data. \textbf{Hybrid} combines BM25 and FT as described in Section~\ref{model_fine_tuning}. Bold font indicates top values w.r.t. the first 3 and last 4 rows.

% \begin{table*}[]
% % \addtolength{\tabcolsep}{-0.3em}
% \begin{tabular}{rccccccccccccccccccccc}
% \toprule
%  & ara & dan & deu & eng & fas & fra & hin & ind & ita & jpn & kor & nld & pol & por & rus & spa & swe & tur & vie & zho \\ \midrule
% BM25 & 30.13 & 40.85 & 25.10 & 24.43 & 30.86 & 28.19 & 34.13 & 35.61 & 30.29 & 29.19 & 30.15 & 29.53 & 26.30 & 32.43 & 20.79 & 29.33 & 30.04 & 30.01 & 34.40 & 35.83 \\
% Hybrid & 32.44 & 36.46 & 29.35 & 28.56 & 33.42 & 32.01 & 36.27 & 37.37 & 33.24 & 32.04 & 31.81 & 33.40 & 29.37 & 34.72 & 24.43 & 32.47 & 33.32 & 33.58 & 36.44 & 37.97 \\ \midrule
% mGTE (base) & 71.06 & 74.72 & 58.08 & 60.21 & 63.83 & 64.97 & 75.90 & 76.27 & 69.17 & 61.79 & 75.46 & 63.64 & 65.90 & 70.71 & 58.53 & 66.96 & 70.19 & 70.51 & 75.89 & 84.85 \\
% mE5 (large) & 79.97 & 82.59 & 70.88 & 67.10 & 74.93 & 72.96 & 82.48 & 82.05 & 77.63 & 73.20 & 83.35 & 75.98 & 75.69 & 79.42 & 68.42 & 74.92 & 76.89 & 77.13 & 83.36 & 87.57 \\
% Jina (v3) & 85.45 & 87.94 & 74.68 & 67.02 & 79.23 & 77.29 & 86.29 & 85.18 & 82.87 & 75.16 & 87.39 & 80.27 & 81.35 & 84.22 & 72.62 & 78.67 & 83.36 & 82.24 & 86.90 & 89.17 \\ \bottomrule
% \end{tabular}
% \end{table*}

% \vfill

\begin{table*}[h]
\caption{Retrieval performance across models in NDCG@10 (\%) on WebFAQ}
\label{tab:full_results_webfaq}
\vspace{-0.2cm}
\addtolength{\tabcolsep}{-0.135em}
\begin{tabular}{rccccccccccccccccccccc}
\toprule
 & ara & dan & deu & eng & fas & fra & hin & ind & ita & jpn & kor & nld & pol & por & rus & spa & swe & tur & vie & zho \\ \midrule[1.5pt]
mGTE & 69.5 & 75.0 & 57.6 & 58.9 & 63.2 & 63.6 & 76.3 & 75.7 & 68.1 & 61.2 & 74.4 & 63.4 & 66.1 & 70.1 & 57.0 & 66.1 & 70.5 & 59.1 & 76.2 & 84.1 \\
mE5 & 79.2 & 82.5 & 70.1 & 66.2 & 74.1 & 72.0 & 82.5 & 81.9 & 76.9 & 72.7 & 82.7 & 76.0 & 75.5 & 78.7 & 66.8 & 74.3 & 77.2 & 65.8 & 83.7 & 87.4 \\
Jina & \textbf{84.6} & \textbf{87.5} & \textbf{74.2} & \textbf{66.4} & \textbf{78.7} & \textbf{76.5} & \textbf{86.6} & \textbf{84.8} & \textbf{82.1} & \textbf{74.5} & \textbf{85.8} & \textbf{80.1} & \textbf{81.5} & \textbf{83.7} & \textbf{71.4} & \textbf{78.3} & \textbf{82.8} & \textbf{70.6} & \textbf{86.9} & \textbf{88.9} \\ \midrule[1.5pt]
BM25 & 29.7 & 34.0 & 25.1 & 24.4 & 30.1 & 27.4 & 34.3 & 36.1 & 30.4 & 29.0 & 29.6 & 29.7 & 26.7 & 32.7 & 21.2 & 29.0 & 30.6 & 25.9 & 34.4 & 35.7 \\ \midrule
Base & 61.0 & 74.7 & 53.1 & 49.7 & 63.2 & 54.9 & 70.1 & 74.6 & 60.0 & 56.5 & 73.2 & 61.2 & 61.5 & 65.3 & 50.1 & 57.1 & 67.6 & 49.8 & 70.2 & 73.5 \\
FT & \textbf{74.1} & \textbf{83.7} & \textbf{66.3} & \textbf{59.0} & \textbf{74.4} & \textbf{69.1} & \textbf{79.6} & \textbf{82.7} & \textbf{73.8} & \textbf{68.8} & \textbf{81.6} & \textbf{74.6} & \textbf{74.2} & \textbf{78.3} & \textbf{61.5} & \textbf{69.6} & \textbf{78.8} & \textbf{66.1} & \textbf{80.7} & \textbf{84.8} \\ \midrule
Hybrid & 32.4 & 36.5 & 29.4 & 28.6 & 33.4 & 32.0 & 36.3 & 37.4 & 33.2 & 32.0 & 31.8 & 33.4 & 29.4 & 34.7 & 24.4 & 32.5 & 33.3 & 33.6 & 36.4 & 38.0 \\ \bottomrule[1.5pt]
\end{tabular}
\end{table*}

% \vfill

\begin{table*}[h]
\caption{Retrieval performance across models in NDCG@10 (\%) on MIRACL (Hard Negatives)}
\label{tab:full_results_mircal}
\vspace{-0.2cm}
% \addtolength{\tabcolsep}{-0.015em}
\begin{tabular}{rcccccccccccccccccc}
\toprule
 & ara & ben & deu & eng & fas & fin & fra & hin & ind & jpn & kor & rus & spa & swa & tel & tha & yor & zho \\ \midrule[1.5pt]
mGTE & 71.8 & \textbf{72.7} & 51.2 & \textbf{54.9} & 52.2 & 73.5 & \textbf{55.2} & 52.3 & \textbf{50.5} & \textbf{66.4} & 63.9 & 63.9 & \textbf{53.0} & 69.9 & \textbf{83.1} & \textbf{74.5} & 58.3 & \textbf{62.3} \\
mE5 & \textbf{75.6} & 72.1 & 50.8 & 46.7 & \textbf{56.4} & \textbf{74.6} & 48.8 & \textbf{58.2} & 50.4 & 66.1 & \textbf{65.8} & 63.5 & 49.1 & \textbf{71.7} & 82.6 & 77.2 & \textbf{59.9} & 53.4 \\
Jina & 72.2 & 71.9 & \textbf{53.3} & 52.1 & 53.9 & 71.9 & 54.9 & 56.7 & 49.4 & 66.3 & 64.3 & \textbf{65.4} & 50.7 & 59.4 & 81.8 & 76.0 & 48.1 & 56.9 \\ \midrule[1.5pt]
BM25 & 53.2 & 53.9 & 21.8 & 31.6 & 34.1 & 51.8 & 21.9 & 48.6 & 51.6 & 44.7 & 37.6 & 29.8 & 36.3 & 53.0 & 50.0 & 27.6 & \textbf{64.4} & 33.9 \\ \midrule
Base & 36.1 & 32.0 & 32.8 & 30.6 & 36.2 & 42.1 & 25.9 & 31.9 & 28.1 & 27.6 & 39.3 & 29.6 & 30.2 & 23.1 & 34.6 & 41.0 & 11.3 & 26.9 \\
FT & 49.3 & 46.0 & \textbf{35.8} & 36.9 & \textbf{43.0} & 55.9 & \textbf{39.5} & 37.5 & 36.3 & 38.5 & 41.2 & \textbf{38.5} & 40.1 & 41.4 & \textbf{59.6} & \textbf{52.1} & 28.6 & 36.7 \\ \midrule
Hybrid & \textbf{61.7} & \textbf{61.1} & 30.4 & \textbf{50.8} & 37.9 & \textbf{62.0} & 29.0 & \textbf{54.8} & \textbf{57.8} & \textbf{51.3} & \textbf{43.2} & 35.2 & \textbf{43.8} & \textbf{57.9} & 59.0 & 34.0 & 56.7 & \textbf{40.3} \\ \bottomrule[1.5pt]
\end{tabular}
\end{table*}

% \vfill

\begin{table*}[h]
\caption{Retrieval performance across models in NDCG@10 (\%) on Mr. Tydi}
\label{tab:full_results_mrtydi}
\vspace{-0.2cm}
\begin{tabular}{rccccccccccc}
\toprule
 & ara & ben & eng & fin & ind & jpn & kor & rus & swa & tel & tha \\ \midrule[1.5pt]
mGTE & 73.1 & \textbf{73.6} & \textbf{57.1} & 63.9 & 67.8 & 59.9 & 56.5 & 63.7 & 69.6 & \textbf{89.2} & 72.7 \\
mE5 & \textbf{76.5} & 72.6 & 52.5 & \textbf{66.4} & \textbf{71.2} & \textbf{61.9} & \textbf{59.5} & \textbf{65.4} & \textbf{73.0} & 87.4 & \textbf{77.3} \\
Jina & 71.9 & 71.3 & 55.2 & 63.2 & 69.4 & 59.0 & 55.6 & 62.3 & 61.6 & 87.0 & 74.4 \\ \midrule[1.5pt]
BM25 & 43.4 & 50.2 & 19.9 & 34.0 & 50.5 & 25.0 & 24.8 & 29.7 & 53.9 & 51.7 & 27.9 \\ \midrule
Base & 30.5 & 29.9 & 18.1 & 24.4 & 33.2 & 16.0 & 27.7 & 22.5 & 27.7 & 32.0 & 35.7 \\
FT & 44.2 & 45.2 & 28.7 & 38.7 & 48.4 & 31.1 & \textbf{34.8} & 29.5 & 46.7 & 62.2 & \textbf{48.9} \\ \midrule
Hybrid & \textbf{54.4} & \textbf{51.6} & \textbf{31.7} & \textbf{45.1} & \textbf{58.5} & \textbf{37.8} & 30.4 & \textbf{36.0} & \textbf{60.8} & \textbf{66.4} & 35.7 \\ \bottomrule[1.5pt]
\end{tabular}
\end{table*}

\end{document}